\pdfoutput=1

\documentclass[11pt]{article}

\usepackage[preprint]{acl}

\usepackage{times}
\usepackage{latexsym}

\usepackage[T1]{fontenc}

\usepackage[utf8]{inputenc}

\usepackage{microtype}

\usepackage{inconsolata}

\usepackage{graphicx}
\usepackage{xcolor}
\usepackage{subcaption}
\usepackage{enumitem}
\usepackage{url}
\usepackage{amsmath}
\usepackage{multirow}
\usepackage{multicol}
\usepackage{array}
\usepackage{booktabs}
\usepackage{longtable}
\usepackage{ltcaption}
\usepackage[most]{tcolorbox}
\usepackage{listings}

\usepackage{algorithm}
\usepackage{algpseudocode}
\algrenewcommand\algorithmicrequire{\textbf{Input:}}
\algrenewcommand\algorithmicensure{\textbf{Output:}}
\newcommand{\zblue}[1]{\text{\textcolor{blue}{#1}}}

\title{WebRollback: Enhancing Web Agents with Explicit Rollback Mechanisms}

\author{
Zhisong Zhang\textsuperscript{1}$^{\dag}$, Tianqing Fang\textsuperscript{2}, Kaixin Ma\textsuperscript{2}, Wenhao Yu\textsuperscript{2},\\
\textbf{Hongming Zhang\textsuperscript{2}, Haitao Mi\textsuperscript{2}, Dong Yu\textsuperscript{2}}\\
\textsuperscript{1}City University of Hong Kong \quad
\textsuperscript{2}Tencent AI Lab\\
$^{\dag}$\texttt{zhisong.zhang@cityu.edu.hk}
}

\begin{document}
\maketitle

\begin{abstract}
With recent advancements in large language models, web agents have been greatly improved. However, dealing with complex and dynamic web environments requires more advanced planning and search abilities. Previous studies usually adopt a greedy one-way search strategy, which may struggle to recover from erroneous states. In this work, we enhance web agents with an explicit rollback mechanism, enabling the agent to revert back to a previous state in its navigation trajectory. This mechanism gives models the flexibility to directly control the search process, leading to an effective and efficient web navigation method. We conduct experiments on two live web navigation benchmarks with zero-shot and fine-tuning settings. The results demonstrate the effectiveness of our proposed approach.
\end{abstract}

\section{Introduction}

With the advancement of large language models (LLMs), LLM-powered autonomous agents have demonstrated great potential in solving a wide range of real-world tasks \citep{wang2024survey}, among which web navigation is a typical example, as it requires the agent's ability to interact with the dynamic web environment \citep{yao2022webshop,deng2023mind2web,zhou2024webarena}.

One of the main challenges in developing web agents is dealing with the dynamic and ever-changing nature of real-world web environments. For example, web pages might be removed, or the layouts of a website can be reorganized. Due to these complexities, agents may sometimes be trapped in erroneous states, where well-planned actions may fail to produce expected outcomes. Most web agents adopt a one-way greedy search strategy, which may have difficulties getting out of unexpected or unpromising states. A straightforward solution to alleviate this issue is to explicitly incorporate more effective search algorithms.

\begin{figure}[t]
	\centering
	\includegraphics[width=0.48\textwidth]{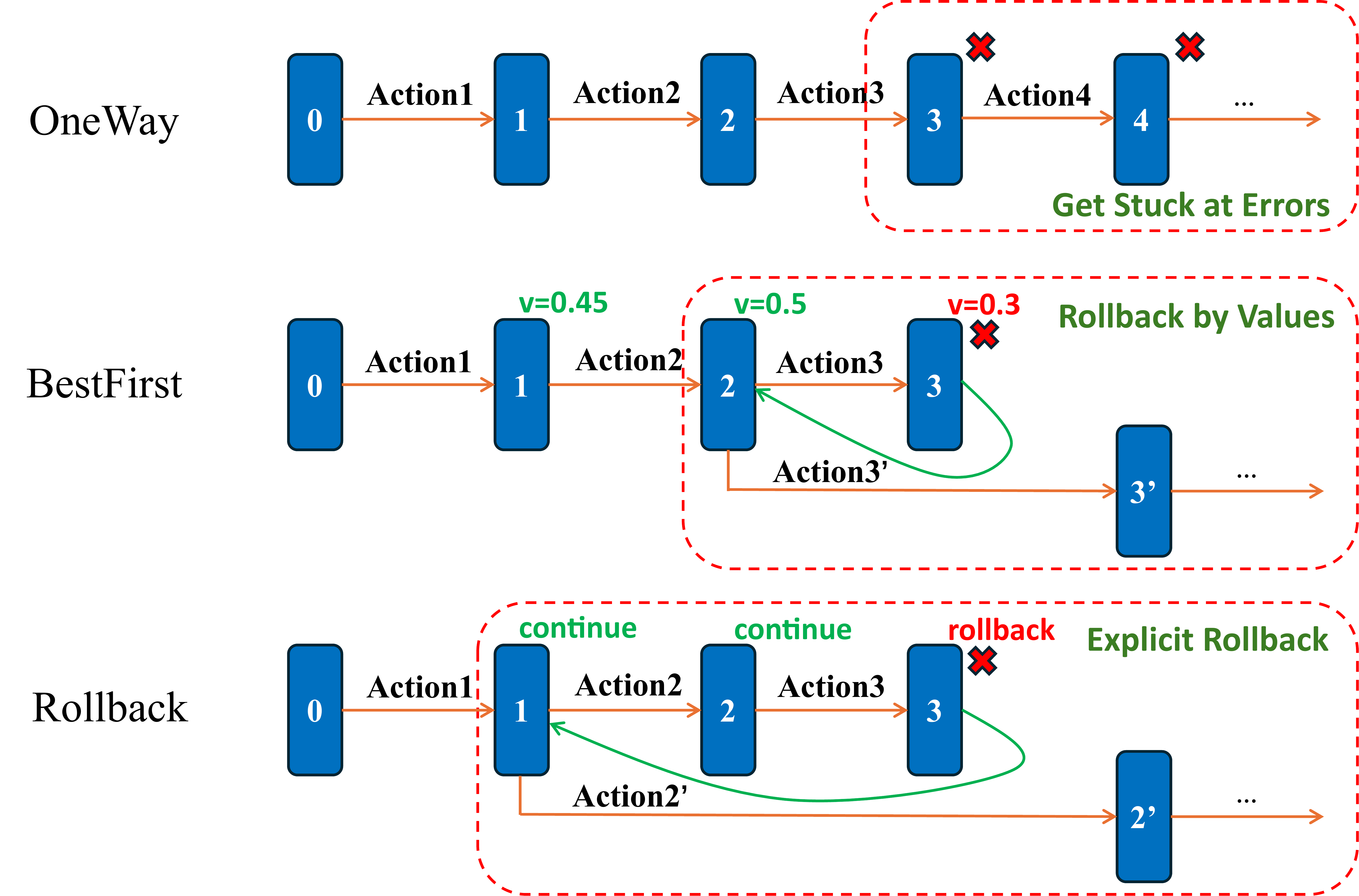}
        \vspace{-0.25in}
	\caption{An overview of different search strategies. The OneWay strategy may get stuck in erroneous states, the BestFirst strategy perform rollback based on state values, while our proposed strategy directly let the models to decide \emph{when and where to rollback}.}
	\label{fig:main}
\end{figure}

Search algorithms have been extensively studied in previous research for LLM reasoning. For non-interactive tasks, parallel search has been shown effective, with common techniques such as majority voting \citep{wang2023selfconsistency}, breadth-first search \citep{yao2023tree}, beam search \citep{xie2023self}, and Monte-Carlo tree search \citep{zhou2024language,chen-etal-2024-tree}. However, it is difficult to directly apply these methods to web navigation tasks considering the much larger cost of maintaining multiple web environments to support parallel interactions. 
Therefore, a suitable search algorithm for web navigation tasks should maintain an overall serialized search process for efficiency while keeping the flexibility to switch to alternative paths to escape from errors. To this end, \citet{koh2024tree} incorporate a best-first search algorithm, which selects the highest-scored state at each step and allows switching to more promising states. However, this approach often results in frequent state switching, which might bring considerable overhead. The main reason is that, for an intermediate state that can lead to the final goal, it may not receive a high score since there is no intermediate evidence of its potential. In this case, the best-first strategy might switch to a seemingly more promising state.

In this work, we propose WebRollback, a strategy that enhances web agents with an explicit rollback mechanism, allowing the agent to revert to a previous state if it judges the current state as erroneous or unpromising. As shown in Figure~\ref{fig:main}, our approach enhances the one-way search algorithm with a rollback mechanism, enabling quick escape from erroneous states while maintaining greater efficiency than the best-first algorithm with fewer state switches. One important feature of our strategy is the flexibility of the rollback operation: \emph{we let the models to more directly influence the search process by deciding when and where to rollback}. Compared with previous backtracking methods\footnote{Many web agents (including ours) already include a ``go-back'' action which supports rolling back a single step.} for reasoning and agent tasks \citep{lutz2024wilbur,yang2025step}, our approach can achieve multi-step rollbacks in a single operation. One contemporary work \citep{li2025generator} similarly investigates a rollback-based approach by identifying the error locations. However, their work primarily addresses agent tasks with clean environments, which are less complex than the real-world web tasks that we explore in this research.

\begin{figure}[t]
	\centering
	\includegraphics[width=0.48\textwidth]{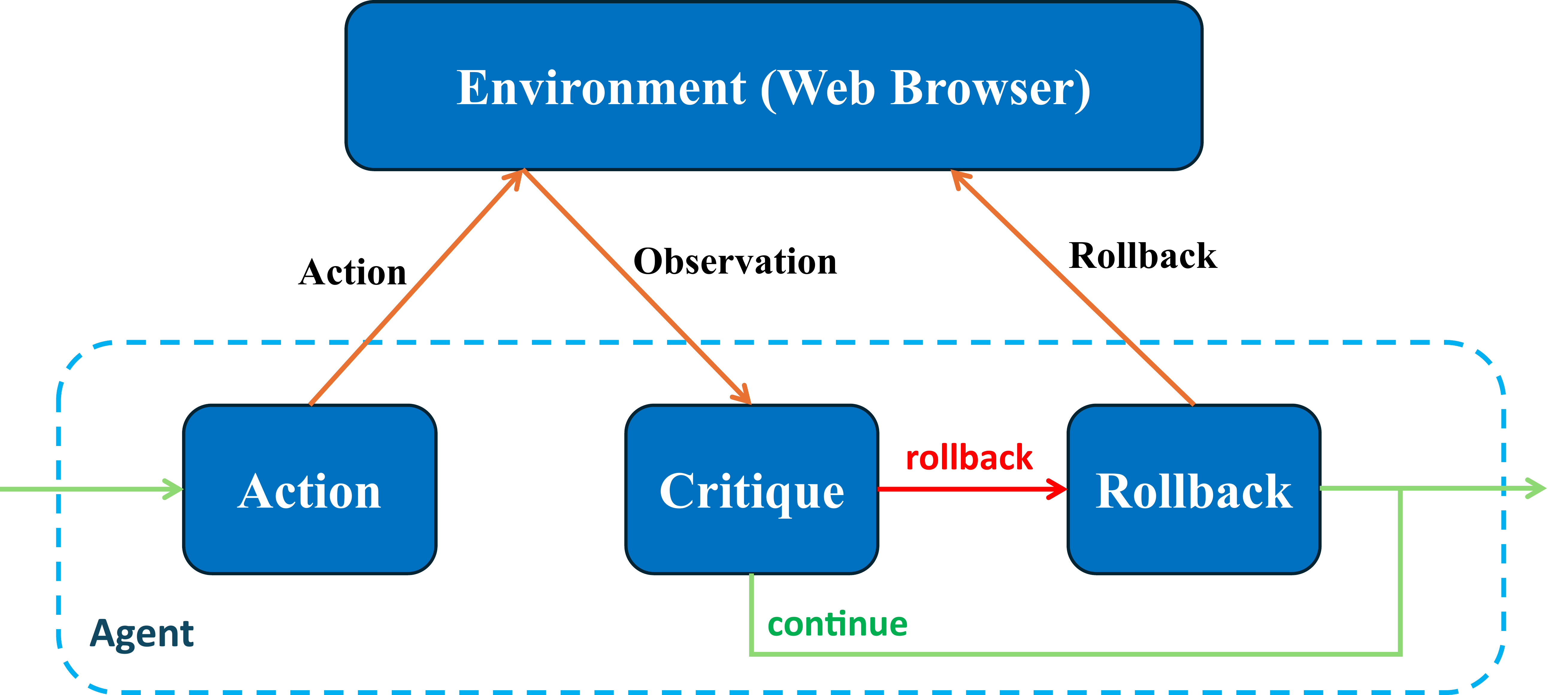}
	\caption{An illustration of the components in our web agent enhanced with explicit rollback mechanisms.}
	\label{fig:agent}
\end{figure}

We conduct experiments with two typical settings: zero-shot prompting with strong LLMs and fine-tuning relatively smaller LLMs using demonstration data from larger models. Through evaluations on two real-world web navigation benchmarks, we show the effectiveness of our proposed approach, which can achieve the best overall results while maintaining efficiency.

\section{Method}


Given a web navigation task, the agent needs to interact with a web browser to solve the task in a step-by-step way. The agent starts with an initial state $s_0$, where the browser is on the homepage of the target website. At each step $i$, the agent receives an observation $o_i$ (i.e., the current webpage) from the environment at the current state $s_i$. Based on this observation, the agent performs an appropriate action $a_i$. This action transitions the environment to a new state $s_{i+1}$, which in turn provides a new observation $o_{i+1}$ to the agent for the next decision. These iterations continue until the target task is resolved or the navigation budget is exhausted.


\begin{algorithm}[t]
    \caption{Search with Rollback.}
	\label{algo:r}
	\begin{algorithmic}[1]
		\small
		\Require Web Environment $env$, Maximum Step $T$.
        \State $S ~\leftarrow~ []$  \Comment{\zblue{Navigation trajectory}}
        \For{$t$ in \text{range}(T)}
            \State $a ~\leftarrow~ \textsc{Action}(env.state)$ \Comment{\zblue{Decide next action}}
            \State \textbf{if} $a$ == ``stop'' \textbf{then} \textbf{break} \textbf{end if} \Comment{\zblue{Task finished}}
            \State $env.\text{step}(a)$ \Comment{\zblue{Execute the action}}
            \State $c ~\leftarrow~ \textsc{Critique}(a, env.state)$ \Comment{\zblue{When to rollback}}
            \If{$c$ == ``rollback''}
            \State $i ~\leftarrow~ \textsc{Rollback}(a, c, S)$ \Comment{\zblue{Where to rollback}}
            \State $env.\text{rollback}(S[i])$ \Comment{\zblue{Perform rollback}}
            \State $S ~\leftarrow~ S[:i+1]$ \Comment{\zblue{Slice trajectory}}
            \Else
            \State $S.\text{append}(env.state)$ \Comment{\zblue{Continue with current}}
            \EndIf
        \EndFor
	\end{algorithmic}
\end{algorithm}


\begin{table*}[t]
	\centering
	\footnotesize
	\begin{tabular}{l | c c c c | c c c c}
		\toprule
        & \multicolumn{4}{c|}{Mind2WebLive} & \multicolumn{4}{c}{Webvoyager} \\
        \midrule
        & Full\%($\uparrow$) & Partial\%($\uparrow$) & Step($\downarrow$) & Switch($\downarrow$) & Full\%($\uparrow$) & Partial\%($\uparrow$) & Step($\downarrow$) & Switch($\downarrow$) \\
        \midrule
        \multicolumn{9}{l}{\emph{Llama3.3-70B-Instruct}} \\
        \midrule
OneWay & $20.92_{\pm 3.61}$ & $41.59_{\pm 2.75}$ & $14.4_{\pm 0.2}$ & $0$ & $38.06_{\pm 1.70}$ & $57.55_{\pm 2.20}$ & $12.2_{\pm 0.2}$ & $0$ \\
BestFirst & $21.16_{\pm 0.41}$ & $\mathbf{46.45}_{\pm 0.41}$ & $14.7_{\pm 0.2}$ & $8.1_{\pm 0.1}$ & $39.82_{\pm 3.16}$ & $61.47_{\pm 3.18}$ & $12.0_{\pm 0.1}$ & $5.3_{\pm 0.1}$ \\
Rollback & $\mathbf{24.07}_{\pm 1.42}$ & $45.29_{\pm 0.20}$ & $\mathbf{14.1}_{\pm 0.2}$ & $5.0_{\pm 0.3}$ & $\mathbf{44.30}_{\pm 1.32}$ & $\mathbf{63.24}_{\pm 2.23}$ & $\mathbf{11.8}_{\pm 0.2}$ & $3.2_{\pm 0.2}$ \\
        \midrule
        \multicolumn{9}{l}{\emph{Qwen2.5-72B-Instruct}} \\
        \midrule
OneWay & $24.53_{\pm 0.70}$ & $43.49_{\pm 0.85}$ & $\mathbf{13.0}_{\pm 0.1}$ & $0$ & $49.56_{\pm 1.61}$ & $65.21_{\pm 2.54}$ & $\mathbf{10.2}_{\pm 0.3}$ & $0$ \\
BestFirst & $23.82_{\pm 1.18}$ & $\mathbf{48.84}_{\pm 2.30}$ & $13.6_{\pm 0.1}$ & $6.3_{\pm 0.1}$ & $47.95_{\pm 2.34}$ & $64.84_{\pm 3.30}$ & $10.8_{\pm 0.1}$ & $4.3_{\pm 0.2}$ \\
Rollback & $\mathbf{27.36}_{\pm 0.94}$ & $45.57_{\pm 1.42}$ & $13.5_{\pm 0.2}$ & $4.5_{\pm 0.1}$ & $\mathbf{51.90}_{\pm 4.24}$ & $\mathbf{69.12}_{\pm 3.79}$ & $\mathbf{10.2}_{\pm 0.1}$ & $2.1_{\pm 0.2}$ \\
		\bottomrule
	\end{tabular}
        \vspace{-0.1in}
	\caption{Zero-shot results with \textsc{Llama-3.3-70B-Instruct} and \textsc{Qwen2.5-72B-Instruct}.}
    \label{tab:zero_shot}
\end{table*}

We enhance the search process of web agents with an explicit rollback mechanism by incorporating model-based rollback modules.
Algorithm~\ref{algo:r} provides an outline of our strategy.
Similar to the greedy OneWay method, our strategy first predicts and executes an action based on the current observation (lines 3-5). Next, a binary decision is made to either continue or roll back (line 6). If a rollback is selected, the rollback operation\footnote{Specifically, if the agent decides to roll back, it outputs the index of the step to return to. Then the environment rewinds to the state corresponding to that index.} will be performed by resetting the browser environment and the state trajectory (lines 7-10). In addition to the action module that generates actions based on the current observation, we introduce two additional components to facilitate the rollback mechanism: a critique module and a rollback module.

The critique module first makes judgments on the new state after executing the previous action, inspired by the self-refinement strategies for LLM reasoning \citep{shinn2023reflexion,madaan2023self}. In addition to the judgments, the module also makes a binary decision: either to proceed with the current state or to revert to a previous one. This approach allows the model to directly determine \emph{when to rollback}, offering a flexible and straightforward method to make such decisions.

Upon a ``rollback'' decision, an explicit rollback module is used to decide \emph{where to rollback}. We list all preceding states as inputs, from which the module selects a specific state to revert to. This strategy provides a flexible and efficient way to perform multi-step rollbacks, which can help the agent quickly escape erroneous states.

\section{Experiments}

\subsection{Settings}

In our agent framework, we use open-source LLMs as the backbone for all the modules. We carefully design prompts for the modules to realize their corresponding functionalities; detailed prompts can be found in Appendix~\ref{app:prompts_agent}. We adopt a text-based benchmarking environment that represents web pages with accessibility trees, and the automatic browser is implemented using Playwright.\footnote{\url{https://playwright.dev/}} For the rollback mechanism, we record the URL of each step and reset the browser to the corresponding URL if a rollback is triggered. Unless specified otherwise, we set a maximum step budget of 16 and stop the navigation if this budget is used up.


\begin{figure}[t]
	\centering
	\includegraphics[width=0.475\textwidth]{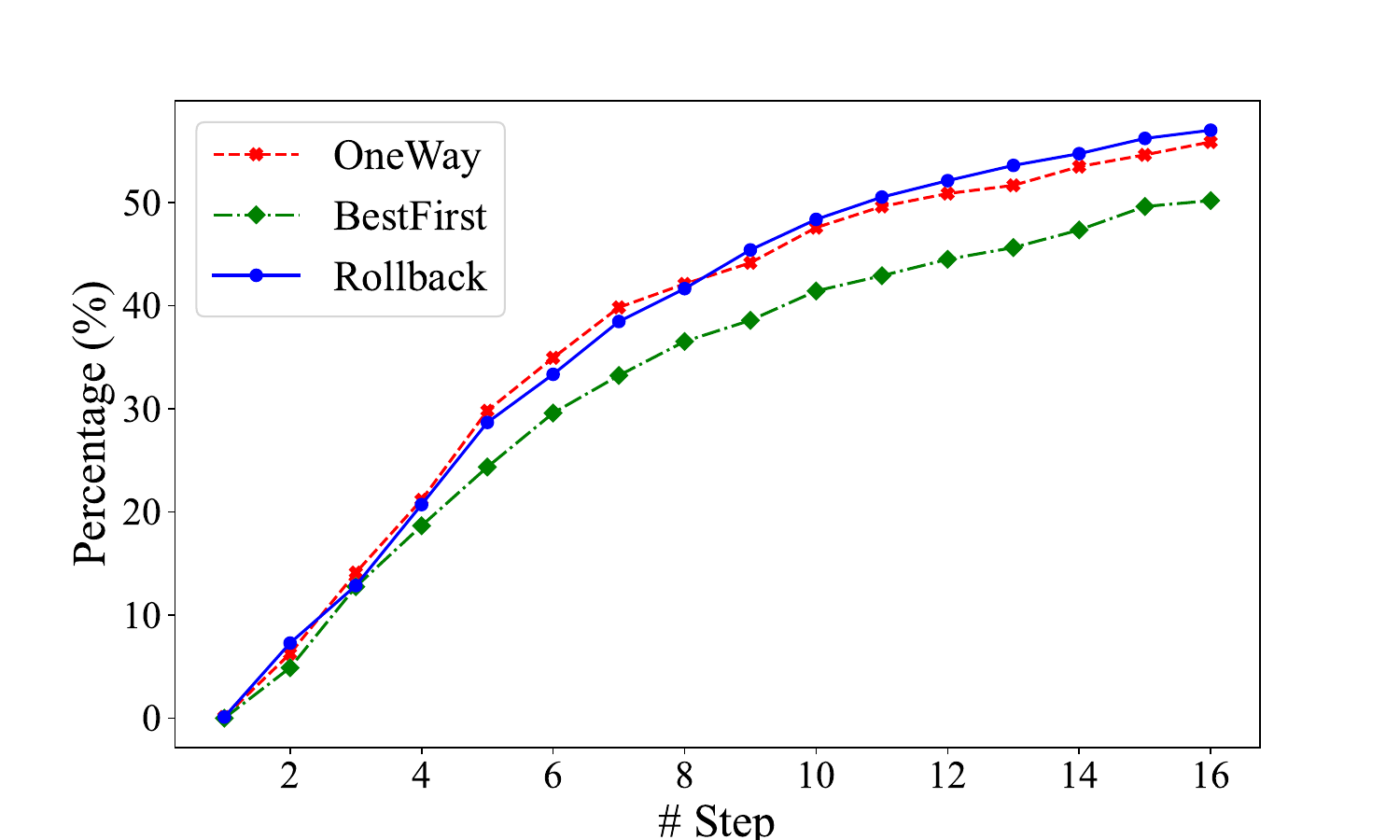}
	\caption{Task-finishing rate analysis. Here, $x$-axis denotes the number of steps the agent takes, and the $y$-axis denotes the percentage of the tasks that can be finished within a specific step limit.}
	\label{fig:ana1}
\end{figure}

We mainly compare three search strategies:\footnote{In our preliminary experiments, we also tried other methods, such as beam-search, but found them to be much less effective and efficient than the above three strategies.} 1) OneWay, which is a greedy-styled search strategy that does not perform any explicit rollbacks except the built-in ``go-back'' action, 2) BestFirst \cite{koh2024tree}, which applies state switches according to the critique's value estimation, and 3) Rollback, which is our strategy with the enhancement of specified rollback modules.

We adopt two live web navigation benchmarks for our evaluation -- Mind2Web-Live \citep{pan2024webcanvas} and WebVoyager \citep{he-etal-2024-webvoyager} -- that involve interacting with real-world websites. These benchmarks are effective for assessing web agents' abilities to handle real-world environments, where unexpected problems or errors may occur. To ensure robustness, we conduct our experiments three times and report the average results. We use GPT-4o as our automatic evaluator, allowing more comprehensive assessments of the navigation trajectory. The scorer assigns fine-grained scores between 0.0 and 1.0 for the successful completion of the target task. The evaluation prompt is provided in Appendix~\ref{app:prompts_eval} and more details about the experimental settings can be found in Appendix~\ref{app:settings}.

\subsection{Zero-shot}

\begin{table*}[t]
	\centering
	\footnotesize
	\begin{tabular}{l | c c c c | c c c c}
		\toprule
        & \multicolumn{4}{c|}{Mind2WebLive} & \multicolumn{4}{c}{Webvoyager} \\
        \midrule
        & Full\%($\uparrow$) & Partial\%($\uparrow$) & Step($\downarrow$) & Switch($\downarrow$) & Full\%($\uparrow$) & Partial\%($\uparrow$) & Step($\downarrow$) & Switch($\downarrow$) \\
        \midrule
        \multicolumn{9}{l}{\emph{Llama3.1-8B-Instruct}} \\
        \midrule
        OneWay & $20.75_{\pm 3.77}$ & $38.49_{\pm 2.64}$ & $14.5_{\pm 0.1}$ & $0$ & $33.92_{\pm 1.75}$ & $52.97_{\pm 1.68}$ & $\mathbf{11.3}_{\pm 0.1}$ & $0$\\
        BestFirst & $19.81_{\pm 0.94}$ & $38.96_{\pm 1.42}$ & $14.8_{\pm 0.2}$ & $6.1_{\pm 0.2}$ & $32.31_{\pm 0.44}$ & $54.59_{\pm 0.68}$ & $11.9_{\pm 0.1}$ & $4.6_{\pm 0.1}$\\
        Rollback & $\mathbf{21.70}_{\pm 2.83}$ & $\mathbf{41.75}_{\pm 2.50}$ & $\mathbf{14.2}_{\pm 0.1}$ & $4.6_{\pm 0.2}$ & $\mathbf{37.43}_{\pm 1.17}$ & $\mathbf{57.29}_{\pm 2.90}$ & $\mathbf{11.3}_{\pm 0.1}$ & $2.7_{\pm 0.3}$\\
        \midrule
        \multicolumn{9}{l}{\emph{Qwen2.5-7B-Instruct}} \\
        \midrule
        OneWay & $13.38_{\pm 1.57}$ & $31.42_{\pm 1.66}$ & $\mathbf{13.7}_{\pm 0.2}$ & $0$ & $28.78_{\pm 2.26}$ & $47.98_{\pm 1.69}$ & $\mathbf{10.9}_{\pm 0.1}$ & $0$\\
        BestFirst & $17.91_{\pm 2.05}$ & $\mathbf{37.86}_{\pm 1.86}$ & $15.0_{\pm 0.2}$ & $6.7_{\pm 0.2}$ & $30.22_{\pm 1.21}$ & $51.92_{\pm 1.41}$ & $12.1_{\pm 0.1}$ & $4.7_{\pm 0.1}$\\
        Rollback & $\mathbf{20.75}_{\pm 2.67}$ & $37.74_{\pm 2.04}$ & $14.5_{\pm 0.4}$ & $5.4_{\pm 0.1}$ & $\mathbf{33.82}_{\pm 0.60}$ & $\mathbf{54.57}_{\pm 1.63}$ & $11.3_{\pm 0.2}$ & $3.1_{\pm 0.2}$\\
		\bottomrule
	\end{tabular}
	\caption{Fine-tuning results with \textsc{Llama-3.1-8B-Instruct} and \textsc{Qwen2.5-7B-Instruct}.}
    \label{tab:fine_tune}
\end{table*}

\paragraph{Setting.} We start with zero-shot prompting settings, where instruction-tuned LLMs are directly prompted to perform target tasks. We adopt two widely-adopted open-source LLMs\footnote{Appendix~\ref{app:extra} provides additional zero-shot evaluation results with smaller LMs.} for this setting: \textsc{Llama-3.3-70B-Instruct} and \textsc{Qwen2.5-72B-Instruct}, considering their impressive instruction-following abilities.

\paragraph{Result.} The zero-shot results are presented in Table \ref{tab:zero_shot}. Here, ``Full'' and ``Partial'' indicate full task success and fine-grained partial scores, respectively. ``Step'' indicates the average number of navigation steps, while ``Switch'' denotes the number of state switches,\footnote{By design, OneWay does not perform any rollbacks.} where rollbacks are performed. The results demonstrate a consistent pattern: Compared with OneWay, our strategy achieves better results; compared to BestFirst, our strategy requires fewer state switches and performs better in complete task success. These illustrate the effectiveness of our proposed rollback mechanisms.

\paragraph{Analysis.} We further provide a task-finishing rate analysis with different search methods, and the results are shown in Figure~\ref{fig:ana1}. Overall, we can see that our strategy can complete tasks in a similar rate to OneWay. In contrast, BestFirst is less efficient, as it more frequently makes state switches.

\begin{table}[t]
	\centering
	\scriptsize
	\begin{tabular}{l | c c c c}
		\toprule
        & Full\%($\uparrow$) & Partial\%($\uparrow$) & Step($\downarrow$) & Switch($\downarrow$) \\
        \midrule
        \multicolumn{5}{l}{\textsc{GPT-4o-mini}} \\
        \midrule
OneWay & $29.50_{\pm 0.50}$ & $49.57_{\pm 0.62}$ & $13.0_{\pm 0.1}$ & $0$\\
BestFirst & $28.15_{\pm 1.15}$ & $51.19_{\pm 1.44}$ & $13.2_{\pm 0.1}$ & $7.8_{\pm 0.3}$\\
Rollback & $34.85_{\pm 3.54}$ & $56.82_{\pm 2.32}$ & $12.8_{\pm 0.1}$ & $6.3_{\pm 0.3}$\\
        \midrule
        \multicolumn{5}{l}{\textsc{GPT-4o}} \\
        \midrule
OneWay & $38.50_{\pm 0.50}$ & $55.17_{\pm 1.57}$ & $11.5_{\pm 0.1}$ & $0$\\
BestFirst & $32.83_{\pm 1.52}$ & $55.73_{\pm 0.93}$ & $12.4_{\pm 0.1}$ & $7.0_{\pm 0.2}$\\
Rollback & $43.50_{\pm 1.50}$ & $60.00_{\pm 1.30}$ & $11.6_{\pm 0.2}$ & $3.5_{\pm 0.3}$\\
		\bottomrule
	\end{tabular}
	\caption{Zero-shot results (on a 100 randomly selected subset) with \textsc{GPT-4o-MINI} and \textsc{GPT-4o}. The overall trends are similar to those with open-source models.}
    \label{tab:zero_shot_gpt}
\end{table}

\paragraph{With Close-Source Models.} We further include results (on a 100 randomly selected subset) with two close-source models: \textsc{GPT-4o-MINI} and \textsc{GPT-4o}. The results are shown in Table \ref{tab:zero_shot_gpt} and the general trends are similar to those in Table \ref{tab:zero_shot}.

\subsection{Fine-tuning}

\paragraph{Setting.} LLMs can achieve impressive results, but usually cost too much; it would be great if we could adopt smaller LMs to complete web navigation tasks. However, when smaller LMs are directly prompted, they often struggle to comprehend complex task instructions, as demonstrated by their poorer performance shown in Table~\ref{tab:zero_shot2} of Appendix~\ref{app:extra}. To address this issue, we seek to enhance smaller LMs' web navigation abilities through knowledge distillation from larger LMs. We use \textsc{Qwen2.5-72B-Instruct} as the teacher model considering its overall better performance, and fine-tune \textsc{Llama-3.1-8B-Instruct} and \textsc{Qwen2.5-7B-Instruct}. The tuning datasets are taken from the training split of Mind2Web-Live and the imitation learning training data from OpenWebVoyager \citep{he2024openwebvoyager}.

\paragraph{Result.} The fine-tuning results are presented in Table~\ref{tab:fine_tune}. The overall patterns are similar to those in the zero-shot results of the teacher models: our proposed approach can achieve a good balance between effectiveness and efficiency, yielding the overall best results at a reasonable cost.

\begin{figure}[t]
	\centering
	\includegraphics[width=0.475\textwidth]{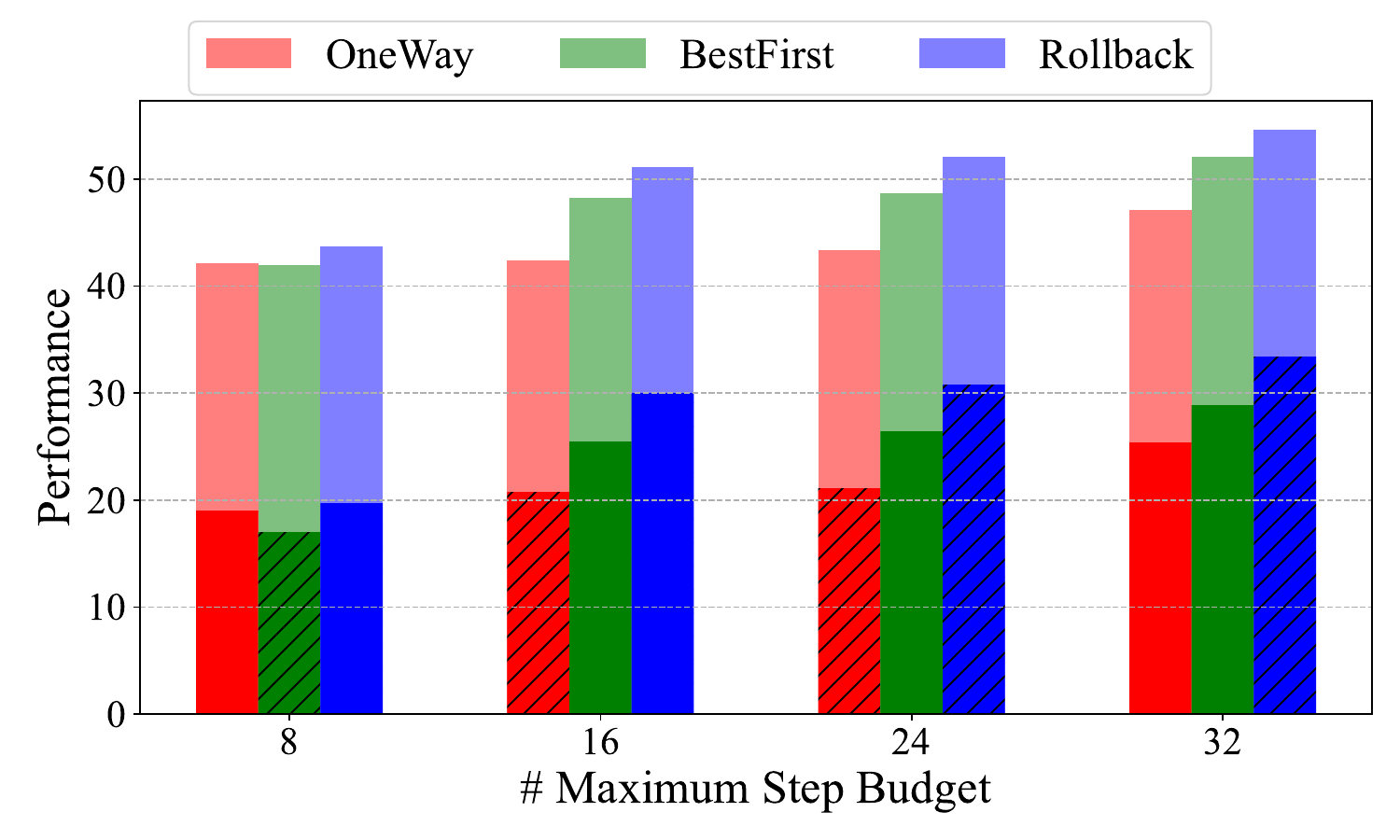}
	\caption{Results with different maximum step budgets. The light bars indicate the Partial\% scores, while the darker and shaded parts represent Full\% scores.}
	\label{fig:anaS}
\end{figure}

\paragraph{Analysis.} We further provide an analysis on test-time scaling \citep{snell2024scaling} by varying the maximum step budget to assess how much the agent can improve with additional search budgets. Due to the higher computational costs of this analysis, we randomly selected 100 instances from the combination of two benchmarks. The results are shown in Figure~\ref{fig:anaS}. When the step budget is limited, such as with 8 steps, different methods perform similarly. However, as the number of steps increases, the strategies with rollback mechanisms improve the results more rapidly than the vanilla OneWay approach. In particular, our proposed strategy consistently achieves the best overall performance, demonstrating its effectiveness in utilizing increased step budgets.

\subsection{Why Explicit Rollback Helps}
Our analyses in Figure \ref{fig:ana1} and \ref{fig:anaS} illustrate that performance improves with increased search budgets, a phenomenon often referred to as test-time scaling \citep{snell2025scaling}. The key question, however, is how to better utilize the additional budgets, for which our method is effective.
For OneWay, the main issue is that it often gets stuck in erroneous states, wasting several steps before recovery (using ``restart'' or consecutive ``goback'' actions). Introducing an explicit multi-step rollback mechanism naturally mitigates this problem. To quantify this effect, we calculate a metric of struggle-ratio, defined as the proportion of instances requiring recovery. We find that OneWay struggles much more (19\% vs 7\%) compared to RollBack.
For BestFirst, we observe frequent state switching because it is designed to choose the highest-valued state. This can lead to unnecessary rollbacks, particularly considering that future rewards are difficult to estimate. In contrast, our RollBack strategy explicitly determines when and where to rollback, providing better controls. The reduced number of state switches confirms this benefit: with fewer unnecessary switches, our method can utilize search budgets in a more effective way.

\section{Conclusion and Discussion}
In this work, we introduce a framework for enhancing web agents with explicit rollback mechanisms, enabling them to autonomously revert to previous states when necessary. Our approach allows the agent to directly decide when and where to rollback, resulting in a flexible and efficient scheme. Experimental results across two web navigation benchmarks demonstrate the effectiveness of our approach in both zero-shot and fine-tuning settings.

While this study focuses on the decision logic of rollbacks, the technical execution of the rollback (how to rollback) presents its own challenges. Specifically, irreversible actions such as processing payments or sending emails require specialized handling. Future research should explore ``high-stakes'' action modeling, where agents adopt greater caution and maintain richer state representations before committing to non-revertible transitions. These directions are vital for deploying reliable agents in real-world applications.

\section*{Limitations}

This work has some limitations. First, we focus solely on web navigation tasks, while the proposed approach could be extended to a wider range of agent tasks. Moreover, due to computational resource constraints, we only tune LMs with a scale of around 7B, while tuning and improving larger LMs could have the potential to achieve better results. Finally, our rollback mechanism is based on a straightforward URL redirection method. However, in many real-world scenarios, reverting actions can be significantly more challenging or even unfeasible due to irreversible changes.

\section*{Ethics Statement}

Our research focuses mainly on the advancement of algorithms to improve web agents. Therefore, we have not applied any additional aggressive filtering techniques to the text data. We use open-source language models in their existing form, without making further modifications to improve safety or minimize bias. As a result, the text data and models that we use may have issues related to offensiveness, toxicity, fairness, or bias that we have not specifically addressed, as these are not the primary objectives of our study. Beyond these points, we do not anticipate any other ethical concerns or risks associated with our research.

\bibliography{main}

\appendix
\onecolumn

\section{Web Agent Prompts}
\label{app:prompts_agent}

\begin{tcolorbox}[colback=blue!5!white, colframe=blue!75!black, breakable, title = {\textsc{Action Module}}]

\ttfamily
\small

You are an assistant helping to browse and operate web pages to solve a specific task.\\

\textbf{Available Information}
\begin{itemize}[leftmargin=*]
    \vspace{-3mm}
    \item Target Task: The specific task you need to accomplish.
    \vspace{-3mm}
    \item Previous Trace Review: A concise review of previous actions and navigation trajectory (before the current state).
    \vspace{-3mm}
    \item Experience: Summaries of previous reverting points, some of which indicate failures. These can guide the next actions by avoiding unsuccessful attempts.
    \vspace{-3mm}
    \item Previous Tryings: Previously tried actions from the current state. Avoid repeating the unsuccessful or already-tried operations.
    \vspace{-3mm}
    \item Accessibility Tree: A simplified representation of the current webpage (web page's accessibility tree), showing key elements in the current window.
\end{itemize}

\textbf{Actions}
\begin{itemize}[leftmargin=*]
    \vspace{-3mm}
    \item click [id] {link name}: Click on an element with a specific `id` on the webpage. Only click on clickable elements like links and buttons.
    \vspace{-3mm}
    \item type [id] {context}: Type the `context` into the field with `id` (this action includes pressing enter by default).
    \vspace{-3mm}
    \item scroll up or scroll down: Scroll the page up or down.
    \vspace{-3mm}
    \item wait: Wait for the page to load (5 seconds).
    \vspace{-3mm}
    \item goback: Navigate to the previously viewed page.
    \vspace{-3mm}
    \item restart: Navigate to the starting URL. Use this if you think you get stuck.
    \vspace{-3mm}
    \item stop [answer] (summary): Issue this action when you believe the task is complete and provide the `answer`. If the task is impossible to complete, provide the answer as "N/A". Include a short `summary` of all previous steps (navigation history) leading to this answer in parentheses.
\end{itemize}

\textbf{Guidelines}
\begin{itemize}[leftmargin=*]
    \vspace{-3mm}
    \item For complex tasks requiring multiple reasoning steps, proceed in a well-planned, step-by-step manner.
    \vspace{-3mm}
    \item Only issue actions that are valid based on the current observation (accessibility tree). For example, do NOT type into buttons, do NOT click on StaticText. If there are no suitable elements in the accessibility tree, do NOT fake ones and do NOT use placeholders like `[id]`.
    \vspace{-3mm}
    \item Issue only one action at a time.
    \vspace{-3mm}
    \item Avoid repeating the same action if the webpage remains unchanged. Maybe the wrong web element or numerical label has been selected. Continuous use of the `wait` action is not allowed.
    \vspace{-3mm}
    \item Issue the `stop` action when the objective has been achieved.
    \vspace{-3mm}
    \item If there is a cookie banner on the page, accept it.
    \vspace{-3mm}
    \item Avoid assuming a specific current date (for example, 2023); use terms like "current" or "latest" if needed. If a specific date is mentioned in the user query, retain that date.
    \vspace{-3mm}
    \item Consider a decomposition-based method that search for simpler queries if a complex query does not yield helpful results.
    \vspace{-3mm}
    \item Remember to try scrolling up or down to find more information on the current page since at each step we only show the accessibility tree restricted to the current window.
    \vspace{-3mm}
    \item Avoid repeated tryings listed in `Previous Tryings` (if existing) since those paths have been tried before. Also check previous experience listed in `Experience` to avoid repeating failures.
\end{itemize}

\textbf{Target Task}\\
\{Target task description\}\\

\textbf{Previous Trace Review}\\
\{Previous trajectories\}\\

\textbf{Experience}\\
\{Experience from previous rollbacks\}\\

\textbf{Previous Tryings}\\
\{Previously tried actions from the current state\}\\

\textbf{Accessibility Tree}\\
\{The accessibility tree for the current web page\}\\

\newpage

\textbf{Output}\\
Please generate your response, your reply should strictly follow the format (each item should be put in one line):
\begin{itemize}[leftmargin=*]
    \vspace{-3mm}
    \item FirstThought: First, provide your brief first thoughts and rationales for your action.
    \vspace{-3mm}
    \item SecondThought: Next, if applicable, examine previous experience (in `Experience` section), previous trace (in `Previous Trace` section) and alternative tryings (in `Previous Tryings` section) from the current state to see if you are repeating previously tried actions. If so, revise your decision and try other options.
    \vspace{-3mm}
    \item Thought: Next, provide a clean version of your final thoughts. In this field, do NOT explicitly mention previous experience (in `Experience` section) and tryings (in `Previous Tryings` section). This field should be used to train models to make good decisions directly. However, still reflect on the steps listed in `Previous Trace` to enhance the model's reflection ability.
    \vspace{-3mm}
    \item Action: Finally, directly output the next action you choose to take. Remember to only issue one action and strictly follow the required formats.
\end{itemize}
\end{tcolorbox}

~
\begin{tcolorbox}[colback=blue!5!white, colframe=blue!75!black, breakable, title = {\textsc{Critique Module}}]

\ttfamily
\small

You are an assistant responsible for evaluating the actions of an intelligent agent navigating a web browser to accomplish specific web-based tasks. Your goal is to assess the agent's latest action and provide constructive feedback.\\

\textbf{Available Information}
\begin{itemize}[leftmargin=*]
    \vspace{-3mm}
    \item Target Task: The specific web-based task the agent aims to complete.
    \vspace{-3mm}
    \item Previous Trace Review: A summary of the agent's past actions and decisions.
    \vspace{-3mm}
    \item Previous Observation: A simplified representation of the previous webpage before the action (previous web page's accessibility tree).
    \vspace{-3mm}
    \item Action to Evaluate: The current step of decision to be evaluated, which transforms the previous web page to the current one.
    \vspace{-3mm}
    \item New Observation: A simplified representation of the new webpage after executing the action (current web page's accessibility tree).
\end{itemize}

\textbf{Evaluator Actions}\\
After evaluation, guide the agent by choosing one of the following actions:
\begin{itemize}[leftmargin=*]
    \vspace{-3mm}
    \item continue: Proceed with the current state.
    \vspace{-3mm}
    \item back: Revert to a previous state.
\end{itemize}

\textbf{Guidelines}
\begin{itemize}[leftmargin=*]
    \vspace{-3mm}
    \item \textbf{Observation and Comparison}: \\
    - Provide brief descriptions of the current accessibility tree.\\
    - Compare the differences between previous and new observations to assess the effects of the agent's actions.
    \vspace{-3mm}
    \item \textbf{Details}: Record important details from the current accessibility tree for future reference.\\
    - Due to space constraints, the accessibility trees are not stored for later steps. Therefore, key information helpful for solving the task should be noted within these fields to avoid loss.\\
    - If there is no useful information in the current webpage, simply fill in "No useful information".
    \vspace{-3mm}
    \item \textbf{Critic}: Evaluate the effectiveness of the current action.\\
    - You are evaluating an intermediate step, which may be partial for the full task. Focus on whether the step and the collected information are helpful towards completing the task, rather than whether the job is finished. The main goal is to correct cases where no or little information is found.\\
    - Carefully check the history to see if the agent is repeatedly performing bad actions. Avoid repeating actions that have already been tried and found unproductive.
    \vspace{-3mm}
    \item \textbf{Scoring Criteria}: Notice that we are only evaluating the current step, which is allowed to be an intermediate and partial step towards the full task.\\
    - score = 1.0: You find that the status of the task is stuck or in an erroneous state, and you need to adjust the direction of your planning and action or revert to a previous state.\\
    - score = 3.0: You find that the current step is reasonable or promising towards completing the target task.\\
    - score = 5.0: You find that the current step is a very critical and successful intermediate step to complete this task.
    \vspace{-3mm}
    \item \textbf{Action}: Decide whether we should continue with the current state or go back to a previous state.\\
    - Notice that the `back` action should be chosen cautiously, which will bring extra costs; it should be used ONLY when you are sure that the current state is not promising for any meaningful progress.\\
    - At many times, you may encounter a state when there are no immediate positive feedback, but if you think it can still lead to promising future states, you are encouraged to try a few more steps (such as scrolling down multiple times).\\
    - However, if you still do not obtain reasonable results after several attempts, then consider reverting to a previous state.
\end{itemize}

\textbf{Target Task}\\
\{Target task description\}\\

\textbf{Previous Trace Review}\\
\{Previous trajectories\}\\

\textbf{Previous Observation}\\
\{The accessibility tree for the previous web page before the action\}\\

\textbf{Action to Evaluate}\\
\{The current step of decision to be evaluated\}\\

\textbf{New Observation}\\
\{The accessibility tree for the new web page after the action\}\\

\textbf{Output}\\
Please generate your response, your reply should strictly follow the format (each item should be put in one line):
\begin{itemize}[leftmargin=*]
    \vspace{-3mm}
    \item Observation: {First, carefully compare the web pages before and after the action (`Previous Observation` and `New Observation`) and briefly describe what are the changes that the action brings.}
    \vspace{-3mm}
    \item Details: {Next, briefly record important information from the current accessibility tree for later reference.}
    \vspace{-3mm}
    \item Critic: {Next, provide your thoughts and evaluations for the action and the current state.}
    \vspace{-3mm}
    \item Score: {Next, rate a float score ranging from 1.0 to 5.0 for the action, in increments of 0.5.}
    \vspace{-3mm}
    \item Action: {Finally, select one action from the options provided: `continue` or `back`.}
\end{itemize}

\end{tcolorbox}

~
\begin{tcolorbox}[colback=blue!5!white, colframe=blue!75!black, breakable, title = {\textsc{Rollback Module}}]

\ttfamily
\small

You are an assistant helping to browse and operate web pages to solve a specific task. In this step, you find that you are stuck or in a bad state and will need to revert back to a previous step. Please read the relevant information and provide your decisions.\\

\textbf{Available Information}
\begin{itemize}[leftmargin=*]
    \vspace{-3mm}
    \item Target Task: The target task you are going to solve.
    \vspace{-3mm}
    \item Previous Trace: All thoughts, actions and observations in previous steps. You should select one from these steps to revert back.
    \vspace{-3mm}
    \item Most Recent Action: The most recent action that you just take.
    \vspace{-3mm}
    \item Critic: Feedback on the latest action, explaining why a revert is necessary.
\end{itemize}

\textbf{Guidelines}
\begin{itemize}[leftmargin=*]
    \vspace{-3mm}
    \item Analyze the current state and previous trace to decide which previous step to revert back.
    \vspace{-3mm}
    \item Provide the index of the previous step to return to, the index of each step can be found in the `Previous Trace` section (annotated with angle brackets "<>").
    \vspace{-3mm}
    \item Summarize the experiences and lessons learned from the navigation trace between the current state and the returning point. This summary should be self-contained and informative for future decisions. (There is no need to describe the learning of the reverting mechanism.)
\end{itemize}

\textbf{Target Task}\\
\{Target task description\}\\

\textbf{Previous Trace}\\
\{A full list of previous trajectories with step indexes\}\\

\textbf{Most Recent Action}\\
\{The most recent action\}\\

\textbf{Critic}\\
\{Feedback on the latest action\}\\

\textbf{Output}\\
Please generate your response, your reply should strictly follow the format (each item should be put in one line):
\begin{itemize}[leftmargin=*]
    \vspace{-3mm}
    \item Analysis: {First, provide your analysis and thoughts for your decisions.}
    \vspace{-3mm}
    \item BackIdx: {Next, Provide the "<index>" of the step to return; you can ONLY go back to the steps annotated with "[Checkpoint Saved]".}
    \vspace{-3mm}
    \item Experience: {Finally, provide a brief, self-contained summary of experiences and lessons learned from the navigation trace between the current state and the returning point.}
\end{itemize}

\end{tcolorbox}

\section{Evaluation Prompt}
\label{app:prompts_eval}

\begin{tcolorbox}[colback=green!5!white, colframe=green!75!black, breakable, title = {\textsc{Evaluation}}]

\ttfamily
\small

You are an assistant tasked with evaluating a web-agent's navigation trace and response to a user's query.\\

As an evaluator, you will be presented with the following primary components to assist you in your role:
\begin{itemize}[leftmargin=*]
    \vspace{-3mm}
    \item Task: A clear and specific directive provided in natural language, detailing the online activity to be carried out. These requirements may include conducting searches, verifying information, comparing prices, checking availability, or any other action relevant to the specified web service (such as Amazon, Apple, ArXiv, BBC News, Booking, etc).
    \vspace{-3mm}
    \item Navigation Trajectory: A series of webpage representations and actions showing the trajectory of performing a web task. It serves as a proof of the actions taken in response to the instruction.
    \vspace{-3mm}
    \item Predicted Response: The predicted textual answer after the navigation process.
    \vspace{-3mm}
    \item Gold Information (Optional): Reference information for your evaluation (such as a successful trajectory or reference response). Sometimes the gold information may be unavailable (N/A).
\end{itemize}
\begin{itemize}[leftmargin=*]
    \vspace{-3mm}
    \item You DO NOT NEED to interact with web pages or perform actions such as booking flights or conducting searches on websites.
    \vspace{-3mm}
    \item You SHOULD NOT make assumptions based on information not presented in the accessibility tree when comparing it to the instructions.
    \vspace{-3mm}
    \item Your primary responsibility is to conduct a thorough assessment of the web task instruction against the outcome depicted in the accessibility tree and in the response, evaluating whether the actions taken align with the given instructions.
    \vspace{-3mm}
    \item NOTE that the instruction may involve more than one task, for example, locating the garage and summarizing the review. Failing to complete either task, such as not providing a summary, should be considered unsuccessful.
    \vspace{-3mm}
    \item NOTE that the accessibility tree is authentic, but the response provided by LLM is generated at the end of web browsing, and there may be discrepancies between the response and the accessibility tree.
    \vspace{-3mm}
    \item Note the difference: 1) Result response may contradict the accessibility tree, then the content of the accessibility tree prevails, 2) The content in the Result response is not mentioned on the accessibility tree, choose to believe the content.
\end{itemize}

Here is the Task:
\{orig\_query\}

Here is the Navigation Trajectory:
\{str\_trajectory\}

Here is the Predicted Response:
\{pred\_answer\}

Here is the Gold Information:
\{gold\_info\}

I'll repeat the target task: "\{orig\_query\}", and the starting URL of the navigation is "\{target\_url\}".\\

Please evaluate the navigation trace for completing this target task. Your output should strictly follow this format for your evaluation:
\begin{itemize}[leftmargin=*]
    \vspace{-3mm}
    \item Summary: {Give a summary of the navigation trajectory.}
    \vspace{-3mm}
    \item Thought: {Provide a brief summary of your thoughts and rationale for the output in one concise line.}
    \vspace{-3mm}
    \item Criteria: {Examine the target task and list important evaluation criteria (crucial sub-steps or information to collect towards completing the target task). Also indicate whether each criterion has been met within the navigation trace and the predicted response.}
    \vspace{-3mm}
    \item Score: {Assign ONE overall float score between 0.0 and 1.0 for the navigation trace and the predicted response. A score of 1.0 indicates full correctness (task completed), 0.0 indicates total failure (no useful information or meaningful sub-steps), and a score in between reflects partial correctness. The score should be assigned based on the criteria analysis. Note that acknowledging failures should not contribute to an increase in the score.}
\end{itemize}

\end{tcolorbox}

\section{Detailed Settings}
\label{app:settings}

We conduct our evaluations using a subset of the testing portion of Mind2Web-Live\footnote{\url{https://huggingface.co/datasets/iMeanAI/Mind2Web-Live/blob/main/mind2web-live_test_20241024.json}} and WebVoyager\footnote{\url{https://github.com/MinorJerry/WebVoyager/blob/main/data/WebVoyager_data.jsonl}}. Due to network connection problems and anti-scraping techniques employed by target websites, we are unable to establish stable connections to certain target websites in the test sets. Therefore, we first perform a connection testing and filter out the websites to which we have trouble connecting. Here is a list of the websites that are excluded:
\begin{tcolorbox}[colback=red!5!white, colframe=red!75!black, breakable, title = {\textsc{Excluded Websites}}]

\ttfamily
\small

EXCLUDED\_WEBSITES = \{
`exploretock', `kohls', `united', `parking', `viator', `delta', `redbox', `soundcloud', `gamestop', `travelzoo', `amctheatres', `ryanair', `cargurus', `resy', `rentalcars', `kbb', `cabelas', `menards', `yellowpages', `tripadvisor', `tiktok.music', `stubhub', `thumbtack', `weather', `uhaul', `health.usnews', `healthgrades', `theweathernetwork', `zocdoc', `usnews.education', `epicurious', `osu.edu', `ups', `dmv.virginia.gov', `extraspace', `finance.yahoo', `pinterest',
`sixflags', `spothero', `justice.gov', `foxsports', `ign', `koa', `tvguide', `webmd', `sports.yahoo', `babycenter', `tesla',
`booking', `dictionary.cambridge.org', `espn', `amazon', `google', `github', `allrecipes',
\}
\end{tcolorbox}
After applying the filtering process, we have 42 and 342 testing instances remaining for Mind2Web-Live and WebVoyager, respectively. For the training datasets for fine-tuning, we adopt the training portion of Mind2Web-Live\footnote{\url{https://huggingface.co/datasets/iMeanAI/Mind2Web-Live/blob/main/mind2web-live_train_20240528.json}} and the imitation learning training data from OpenWebVoyager.\footnote{\url{https://github.com/MinorJerry/OpenWebVoyager/tree/main/WebVoyager/data_for_training/IL}} After applying a similar filtering procedure, the final training set contains a total of 455 instances.

For fine-tuning, we adopt a standard full-model supervised fine-tuning (SFT) recipe using OpenRLHF \citep{hu2024openrlhf}.\footnote{\url{https://github.com/OpenRLHF/OpenRLHF/blob/main/examples/scripts/train_sft_llama.sh}} A single model is trained in a multi-task way to support different modules by using their corresponding prompts.

\section{Extra Results}
\label{app:extra}

\begin{table*}[h]
	\centering
	\footnotesize
	\begin{tabular}{l | c c c c | c c c c}
		\toprule
        & \multicolumn{4}{c|}{Mind2WebLive} & \multicolumn{4}{c}{Webvoyager} \\
        \midrule
        & Full\%($\uparrow$) & Partial\%($\uparrow$) & Step($\downarrow$) & Switch($\downarrow$) & Full\%($\uparrow$) & Partial\%($\uparrow$) & Step($\downarrow$) & Switch($\downarrow$) \\
        \midrule
        \multicolumn{9}{l}{\emph{Llama3.1-8B-Instruct}} \\
        \midrule
OneWay & $0.63_{\pm 0.89}$ & $19.35_{\pm 0.87}$ & $15.9_{\pm 0.1}$ & $0$ &  $3.03_{\pm 1.59}$ & $24.90_{\pm 0.77}$ & $15.9_{\pm 0.1}$ & $0$ \\
BestFirst & $0.63_{\pm 0.89}$ & $22.08_{\pm 1.92}$ & $16.0_{\pm 0.0}$ & $9.1_{\pm 0.5}$ & $3.13_{\pm 0.13}$ & $29.91_{\pm 1.25}$ & $15.9_{\pm 0.1}$ & $6.6_{\pm 0.2}$ \\
Rollback & $0.63_{\pm 0.89}$ & $22.14_{\pm 0.54}$ & $16.0_{\pm 0.0}$ & $8.3_{\pm 0.2}$ & $3.23_{\pm 0.25}$ & $28.66_{\pm 1.55}$ & $15.9_{\pm 0.0}$ & $5.6_{\pm 0.3}$ \\
        \midrule
        \multicolumn{9}{l}{\emph{Qwen2.5-7B-Instruct}} \\
        \midrule
OneWay & $3.77_{\pm 2.67}$ & $11.89_{\pm 2.80}$ & $15.5_{\pm 0.4}$ & $0$ &  $11.79_{\pm 0.60}$ & $22.27_{\pm 0.30}$ & $14.5_{\pm 0.1}$ & $0$\\
BestFirst & $1.26_{\pm 1.78}$ & $12.23_{\pm 1.19}$ & $15.8_{\pm 0.1}$ & $12.5_{\pm 0.7}$ & $13.16_{\pm 1.49}$ & $26.13_{\pm 1.59}$ & $14.5_{\pm 0.1}$ & $9.4_{\pm 0.2}$\\
Rollback & $5.03_{\pm 2.35}$ & $14.91_{\pm 2.89}$ & $15.4_{\pm 0.2}$ & $12.0_{\pm 0.2}$ & $12.98_{\pm 1.36}$ & $26.24_{\pm 1.25}$ & $14.4_{\pm 0.2}$ & $8.5_{\pm 0.1}$\\
		\bottomrule
	\end{tabular}
	\caption{Zero-shot results with \textsc{Llama-3.1-8B-Instruct} and \textsc{Qwen2.5-7B-Instruct}. The performance is much worse with these ``smaller'' LMs if not fine-tuned, when compared to the 70B-level ones.}
    \label{tab:zero_shot2}
\end{table*}

\end{document}